\documentclass[runningheads,a4paper]{llncs}

\usepackage{amssymb}
\usepackage{graphicx}
\usepackage{svg}
\usepackage{multirow}

\usepackage[T2A]{fontenc}
\usepackage[utf8]{inputenc}

\usepackage{url}
\urldef{\mailsa}\path|{anonymous.authors}@springer.com |
\newcommand{\keywords}[1]{\par\addvspace\baselineskip
\noindent\keywordname\enspace\ignorespaces#1}

\begin{document}

\mainmatter  

\title{Combining Neural Language Models for Word Sense Induction}


%
%
\author{Nikolay Arefyev\inst{1,2}
\and Boris Sheludko\inst{1,2}
\and Tatiana Aleksashina\inst{3}}
\authorrunning{Combining Neural Language Models for Word Sense Induction}


\institute{Samsung R\&D Institute Russia, Moscow, Russia\\
\and Lomonosov Moscow State University, Moscow,   Russia
\and SlickJump, Moscow,  Russia}

%
%

\toctitle{Lecture Notes in Computer Science}
\tocauthor{Authors' Instructions}
\maketitle

\begin{abstract}
Word sense induction (WSI) is the problem of grouping occurrences of an ambiguous word according to the expressed sense of this word. Recently a new approach to this task was proposed, which generates possible substitutes for the ambiguous word in a particular context using neural language models, and then clusters sparse bag-of-words vectors built from these substitutes. In this work, we apply this approach to the Russian language and improve it in two ways. First, we propose methods of combining left and right contexts, resulting in better substitutes generated. Second, instead of fixed number of clusters for all ambiguous words we propose a technique for selecting individual number of clusters for each word. Our approach established new state-of-the-art level, improving current best results of WSI for the Russian language on two RUSSE 2018 datasets by a large margin.
\end{abstract}

\section{Introduction}
Ambiguity, including lexical ambiguity, when single word has several meanings, is one of the fundamental properties of natural languages, and is among the most challenging problems for NLP. For instance, modern neural machine translation systems are still surprisingly bad at translating ambiguous words \cite{RUSSE_Arefyev}, although there is some progress in the latest Transformer-based systems compared to previously popular RNN-based ones \cite{self_attention}.
Word Sense Induction (WSI) task can be seen as clustering of occurrences of an ambiguous word according to their meaning. A dataset for WSI consists of text fragments containing ambiguous words. Each occurrence of these words is hand-labeled with the expressed sense according to some sense inventory (a dictionary or a lexical ontology). A WSI system gets a list of ambiguous words and text fragments as an input. For each ambiguous word the system should cluster its occurrences into unknown number of clusters corresponding to this word's senses. This is in contrast to Word Sense Disambiguation (WSD) task where systems are also given the sense inventory used by annotators, so both the number of senses and some contextual information about these senses (in the form of their definitions or related words) are available.

Recently \cite{amrami_goldberg} has proposed a new approach to WSI that generates contextual substitutes (i.e. words, which can appear instead of the ambiguous word in a particular context) using bidirectional neural language models (LMs), and clusters TFIDF-scaled bag-of-words vectors of the substitutes. This approach showed SOTA results in SemEval 2013 WSI shared task for English.  For instance, for the word \textit{build} substitutes like \textit{manufacture}, \textit{make}, \textit{assemble}, \textit{ship}, \textit{export} are generated when it is used in \textit{Manufacturing some goods} sense and \textit{erect}, \textit{rebuild}, \textit{open} are generated for the \textit{Constructing a building} sense which allows distinguishing these senses. We improved this approach in two ways. First, the base approach simply unites substitutes retrieved from probability distributions estimated by forward and backward LMs each given only one-sided context. This results in noisy substitutes when either left or right context is short or non-informative. We explored several methods of combining forward and backward distributions and show that substitutes retrieved from a combined distribution perform much better. Second, the base method used the same number of clusters for each word (average number of senses per word was found to be optimal). We show that using a fixed number of clusters for all words has a huge negative effect on the WSI results and propose a method for selecting individual number of clusters for each ambiguous word.
Our approach has achieved new SOTA results on the RUSSE 2018 WSI shared task for the Russian language \cite{panchenko_2018} with a large improvement over previous best results according to the official leaderboard\footnote{\url{ https://competitions.codalab.org/competitions/public_submissions/17806}, \url{ https://competitions.codalab.org/competitions/public_submissions/17809}, see post-competition tabs}. Also we compare performance of several pretrained Russian neural LMs in WSI.
\section{Related work}
Existing WSI methods can be roughly categorized by their relatedness to one of the following lines of research. 
\textbf{Latent Variable Methods} define a probabilistic model of a text generation process that includes latent variables corresponding to word senses. Posterior probability given unlabeled corpus is estimated to solve WSI task. For instance, \cite{Lau2013} relies on the Hierarchical Dirichlet Process and \cite{bartunov2015breaking} employs the Stick-breaking Process. In \cite{amplayo2019granularity} a rather complicated custom graphical model is proposed which aims at solving the sense granularity problem.
\textbf{Graph Clustering} methods like \cite{veronis2004hyperlex,Hope2013} first build a graph with nodes corresponding to words, and weighted edges representing semantic similarity strength or co-occurrence frequency. Then graph clustering algorithms are applied to split neighbours of an ambiguous word into clusters interpreted as this word's senses.
\textbf{Context clustering} methods represent each occurrence of an ambiguous word as a vector that encodes its context. For example, in \cite{RUSSE_Arefyev,RUSSE_Kutuzov} a weighted average of the context word embeddings is calculated, then a general clustering algorithm is applied. 

Our work is mostly related to the line of research, which exploits \textbf{contextual substitutes} for the ambiguous word to differentiate between its senses. One of the best performing systems at SemEval 2013 generated substitutes using n-gram language models \cite{baskaya}. Later \cite{amrami_goldberg} proposed using neural language models and a few other tweaks, establishing a new SOTA on this dataset. In section 3 we describe their method with slight modifications (adapting it to the RUSSE 2018 task) as our base approach and then propose several improvements. As an alternative to language models, \cite{alagic_leveraging} propose employing context2vec model \cite{melamud_context2vec} to generate substitutes and building from them the average word2vec representation instead of the bag-of-words representation. Context2vec model merges information from left and right context internally, which may result in better substitutes. However, context2vec requires lots of resources to train and is not readily available for many languages, including Russian. In contrast, neural LMs have become a standard resource available for many languages. Thus, in the current work we focus on using pretrained LMs and improving results for WSI by externally combining information from left and right context. In the preliminary experiments we tried using multilingual BERT model \cite{bert}, which also combines left and right context internally and is pretrained on texts in different languages, including Russian. However, this model's vocabulary consists mainly of subwords (frequently occurring pieces of words similar to morphemes). Using BERT in a naive way (similarly to LMs) to generate substitutes results in small subwords generated that perform poorly for WSI. More sophisticated algorithms like \cite{bertmouth} are required to generate whole words with BERT and we leave it for the future work.

Several competitions were organized to compare approaches to WSI. SemEval 2010 task 14 \cite{manandhar} and SemEval 2013 task 13 \cite{jurgens} are the most popular ones for English. For the Russian language RUSSE 2018 competition \cite{panchenko_2018} has been held recently. Three datasets varying in context length and sense granularity have been built for this competition. In this paper we evaluate our methods on these datasets and compare our results with the best results of this competition. RUSSE 2018 requires hard clustering of text fragments containing an ambiguous word, i.e. each example shall appear in one and only one cluster. In this aspect it is similar to SemEval 2010 and, unlike SemEval 2013, which requires soft clustering (i.e. a probability distribution over clusters for each example). Adjusted Rand Index (ARI) was used as a quality measure of a clustering for each word. The weighted average of ARIs across all ambiguous words with weights proportional to the number of examples per word was used as the final metric. The winner of the competition didn't submit a paper, so little is known regarding the best approach, except that it used algebraic operations on word2vec embeddings to identify word senses \cite{panchenko_2018}. However, the 2nd and the 3rd best results on all datasets were achieved by calculating weighted average of word2vec embeddings for context words and clustering them with either agglomerative clustering or affinity propagation \cite{RUSSE_Arefyev,RUSSE_Kutuzov}.

In the post-competition period \cite{RUSSE_Transformer} proposed pretraining the Transformer sequence transduction model \cite{attention} to recover ambiguous words hidden from it's input (an approach similar to the BERT model pretraining proposed later in \cite{bert}) and using outputs as well as hidden states from this model to represent the ambiguous word in a context. They achieved new SOTA on one of the datasets. According to the official post-evaluation leaderboard no other improvements has been achieved on RUSSE 2018 datasets yet (by the time of May 05 2019 and excluding this paper's results).

\section{Approach}
\label{sec:approach}
\subsection{Baselines}
We use the method from \cite{amrami_goldberg} as our baseline, with slight modifications to account for the differences in the datasets. Suppose our examples look like $l\;c\;r$, where $c$ is the target ambiguous word and $l, r$ are its left and right contexts. The original method does the following:
\begin{enumerate}
    \item Use pretrained forward and backward LMs to estimate probabilities for each word $w$ to be a substitute for $c$ given only the left context $P_{fwd}(w|l)$ or only the right context $P_{bwd}(w|r)$. To provide more information to the LMs and bias it towards generating co-hyponyms of the target word, the target word can be included in the context using dynamic symmetric pattern ``T and \_'' / ``\_ and T''. For instance, for the sentence \textit{These apples are sold everywhere} instead of \textit{``These \_''} we pass \textit{``These apples and \_''} to the forward LM, and instead of \textit{``\_ are sold everywhere''} we pass \textit{``\_ and apples are sold everywhere''} to the backward LM. The underscore denotes the position for which the model predicts possible words.
    \item Take top K predictions from the forward and the backward LM independently, renormalize their probabilities so that they sum to one, and sample L substitutes from each distribution, resulting in 2L substitutes. Do it S times. Each of S sets of substitutes is called a representative of the original example. Build TFIDF BOW vectors for all representatives of all examples for a particular ambiguous word. Additionally, all substitutes are lemmatized to get rid of the grammatical bias (LMs can generate only plural or only singular substitutes depending on the grammatical form of the ambiguous word, so these substitutes will never intersect even for the same sense of the word unless they are lemmatized).
    \item Finally, TFIDF BOW vectors are clustered using agglomerative clustering with cosine distance, average linkage and the same number of clusters for all words (we select it on the train set of each RUSSE 2018 dataset separately while \cite{amrami_goldberg} used the average number of senses in the test set of SemEval 2013). The required probability distribution over clusters for each example is estimated from the number of representatives of this example found in each cluster. 
\end{enumerate}

To obtain hard clustering required by RUSSE 2018 we can simply select the most probable cluster for each example (this method is denoted as \textbf{sampling} in our experiments). However, we have found that skipping sampling and using S=1 representative consisting of top K predictions from each LM, while being conceptually simpler and deterministic, also delivers better results (\textbf{base} method). Also we have found that dynamic symmetric patterns (we have simply translated ``T and \_'' to Russian) sometimes help a little but generally hurt a lot for our best models, so we don't use them by default. This is in line with the ablation study from \cite{amrami_goldberg} showing that the patterns are useful for verbs and adjectives but almost useless for nouns, which RUSSE 2018 datasets consist of. We leave integration of the patterns into our best models and experimenting on other datasets containing other parts of speech for the future work.

\subsection{Combining LMs}

During preliminary experiments we have found that using substitutes retrieved from forward and backward LMs independently results in lots of substitutes not related to the target word. For instance, consider the case when the ambiguous word is the first word of a sentence. The forward LM will simply predict all words which can appear as the first word, and these words will be unrelated to the meaning of the target word. Using patterns like ``T and \_'' / ``\_ and T'' improves this situation to some extent, at least the context will always contain the target word. However, we noticed other problems related to these patterns. Often after \textit{and} the model generates not nouns which are meant to be co-hyponyms of the target word but verbs instead. Probably this is related to the agreement in number between the noun and its syntactically related words. For instance, the LM cannot generate coordinated subjects for a singular predicate, so it tries to generate a coordinate clause instead.

To solve these problems we propose taking top K words from a combination of distributions predicted by forward and backward LMs. We experiment with the following combinations.\\
\textbf{Average (avg)}: $(P_{fwd} + P_{bwd}) / 2$ \\
\textbf{Positionally weighted average (pos-weight-avg)}: $\alpha P_{fwd} + (1-\alpha) P_{bwd}$, where $\alpha$ is a function of normalized (divided by the example length) position of the ambiguous word in the example: $$\alpha(pos) = max( min( 0.5, 0.5{\beta^{-1}} pos ), 0.5{\beta^{-1}} (pos-1+2 \beta)) $$
This allows to weigh forward and backward LMs equally when both left and right context is larger than $\beta$ times example length words and underweigh one of them when corresponding context becomes short. $\beta$ is a hyperparameter to be selected.\\
\textbf{Bayesian combination (bayes-comb)}: using Bayes' rule and supposing left and right context are independent given the target word we estimate the probability we are interested in as $$P(w|l,r) = \frac{P(l,r|w) P(w)}{P(l, r)} = \frac{P(l|w) P(r|w) P(w)}{P(l,r)} \propto \frac{P(w|l)P(w|r)}{P(w)} $$ The numerator is estimated using $P_{fwd}$ and $P_{bwd}$, but pretrained LMs that we use don't contain word frequencies in their vocabularies, so we cannot directly estimate the denominator. However, their vocabularies are sorted by frequency, so we can estimate word frequency ranks and approximate the denominator with Zipf distribution : $P(w) \propto 1/(rank(w))^{z}$. So finally we approximate the conditional probability of a substitute given context as: 
$$ \hat{P}(w|l,r) \propto P_{fwd} P_{bwd} (rank(w))^{z} $$
Interestingly, pointwise mutual information (PMI) which is another popular measure of relatedness between a word and a context can be approximated by exactly the same formula, but with different value of $z$:
$$ PMI(w, (l,r)) = \frac{P(w|l,r)}{P(w)} \propto \frac{P(w|l)P(w|r)}{(P(w))^2} $$
$$ \widehat{PMI}(w, (l,r)) \propto P_{fwd} P_{bwd} (rank(w))^{2z} $$
In contrast to conditional probability, PMI discounts frequent words and promotes rare ones. When we select $z$ as a hyperparameter on the train set, effectively we select from the family of relatedness metrics of the form $ P(w|l,r) / (P(w))^k $ one, which is optimal regarding the final evaluation metric (ARI).

\subsection{Clustering}
Following the original method, we exploit agglomerative clustering, but for each word select individual number of clusters. This approach is not only linguistically more plausible than using the same number of clusters for all words, but also resulted in significant improvement of the final results.

Agglomerative clustering has three hyperparameters: the function defining the distance between examples (affinity), the function defining the distance between clusters (linkage), and the number of clusters. Initially each example is put into a separate cluster. At each iteration, two nearest clusters are merged, until the specified number of clusters is reached. We use cosine affinity and average linkage (the distance between clusters is the average cosine distance between their members). For each word we select the number of clusters, which maximizes the silhouette score:
$$\frac{1}{n} \sum_{i=1}^{n} \frac{b_i - a_i}{max(a_i, b_i)}, $$ where $a_i$ is the mean distance from the i-th example to all examples from the same cluster, and $b_i$ is the mean distance from the i-th example to all examples from the nearest different cluster.
We compare this approach to another one, which sets for all words a single number of clusters selected on the train set.

\section{Experiments and results}

\subsection{Datasets and evaluation}

We evaluate our methods on the three datasets from RUSSE 2018 WSI shared task for the Russian language \cite{panchenko_2018} with different sense granularity, context length and number of examples. Both sense inventory and examples for the active-dict dataset were extracted from the Active Dictionary of Russian \cite{apresjan2011active}, which is an explanatory dictionary of the contemporary Russian language. It contains 253 ambiguous words with 3.5 senses per word on average. The bts-rnc dataset contains examples for 81 ambiguous words extracted from the Russian National Corpus\footnote{\url{http://ruscorpora.ru}} and labeled with their senses according to the Large Explanatory Dictionary\footnote{\url{http://gramota.ru/slovari/info/bts}}, has 3.1 senses per word on average. There are more examples per word compared to the active-dict (124 vs. 23 on average) and the examples are twice as long. The wiki-wiki dataset is the smallest one, it contains only 9 ambiguous words with 2.2 senses and 109 examples per word on average. This dataset contains only homonyms (words having several unrelated senses), so the senses are coarse-grained. Each dataset of RUSSE 2018 is split into train, public test and private test parts. We held out the private test parts to compare our best methods with previous SOTA results, and used the train and the public test parts for models and hyperparameters selection. We didn't use wiki-wiki for development and comparison of LM combination methods due to its small size. However, we do compare our best models to previously best results on all three datasets. 

The metric used in RUSSE 2018 competition for the final ranking of participants is Adjusted Rand Index (ARI). However, different clustering evaluation metrics may exhibit different biases. For better comparison of our results to the previous best results we also adopt two complementary metrics used in the SemEval 2010 competition which also requires hard clustering, namely V-measure and paired F-score \cite{manandhar}. The former metric is biased towards a large number of clusters, while the latter metric prefers small number of clusters, so the geometric mean of these two metrics (AVG) proposed by \cite{DBLP:journals/tacl/WangBGZY15} is usually reported. However, all aforementioned metrics are affected both by vectorization and clustering methods along with their hyperparameters which makes it difficult to compare vectorization methods alone. To estimate vectorization quality while abstracting from clustering hyperparameters selection, we propose and exploit maxARI metric, which is maximum possible ARI achieved by agglomerative clustering with all possible hyperparameter values (varying distance metric, linkage and, most importantly, the number of clusters). Clearly, this is an overestimation of possible results for a particular vectorization when using agglomerative clustering because it selects hyperparameters using gold labels. However, we have found this metric very useful for intermediate comparison and selecting hyperparameters of vectorization methods.

\subsection{Comparison of LMs for the Russian Language}
We compare ELMo LMs \cite{elmo_model} trained on three different corpora (Wikipedia, WMT News and Twitter) for the Russian language.\footnote{\url{http://docs.deeppavlov.ai/en/master/intro/pretrained_vectors.html}} Also we try ULMFiT \cite{ulmfit} LM trained on a subset of the Russian Taiga corpus\footnote{\url{https://github.com/mamamot/Russian-ULMFit}} which is available for the Russian language only as a forward LM. To make the comparison fair we used forward and backward LMs separately. Table~\ref{rus_lms} contains main characteristics of these models taken from corresponding web pages.

\begin{table}[]
\centering
\scriptsize
\begin{tabular}{|c|c|c|c|c|}
\hline
model        & corpora size, tokens & vocabulary size, words & perplexity \\ \hline
elmo-news    & 946M                 & 1M                     & 49.9       \\ 
elmo-twitter & 810M                 & 1M                     & 94.1       \\ 
elmo-wiki    & 386M                 & 1M                     & 43.7       \\ 
ulmfit       & 208M                 & 60K                    & 21.98      \\ \hline
\end{tabular}
\caption{LMs for the Russian language}
\label{rus_lms}
\end{table}


\begin{figure}[h]
\centering
\includegraphics[width=1.0\textwidth]{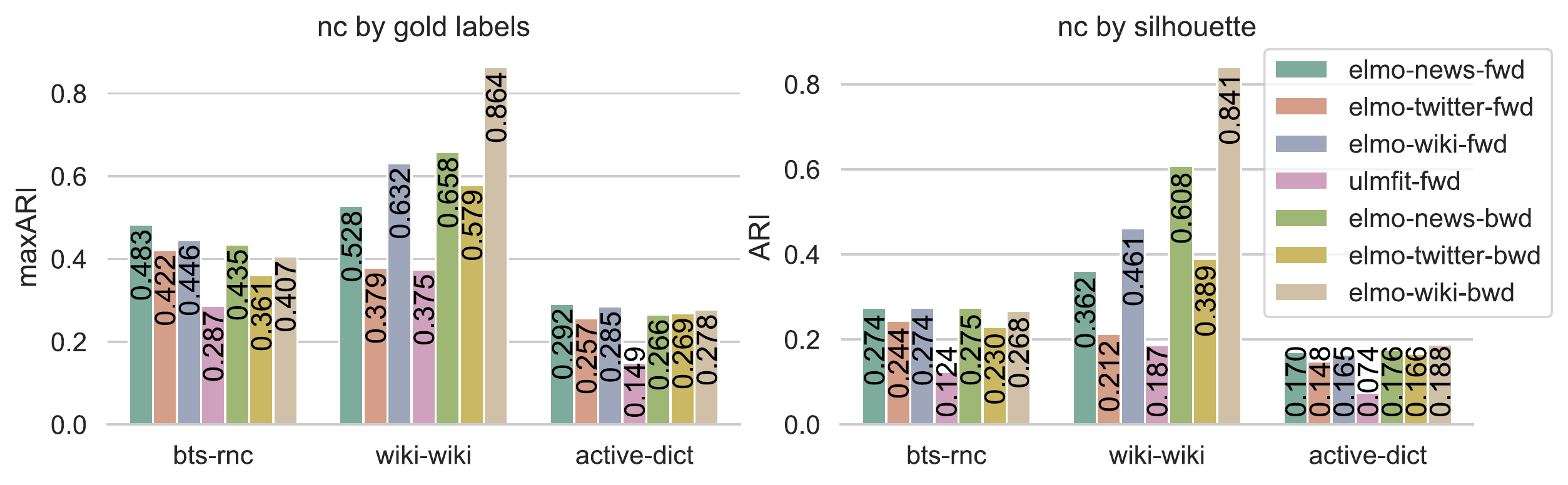}
\caption{Comparison of unidirectional LMs available for Russian. Number of clusters (nc) is selected for each word individually by maximizing either ARI using gold labels (for maxARI estimation) or silhouette score.}
\label{img:cmp_fwd_bwd}
\end{figure}

Figure~\ref{img:cmp_fwd_bwd} shows that ELMo LMs trained on the Russian Wikipedia outperform all other LMs by a large margin on the wiki-wiki dataset, which was also built from the Russian Wikipedia. All backward LMs are significantly better than their forward counterparts on this dataset due to longer right contexts. This is simply because contexts in wiki-wiki are rather large and frequently contain several occurrences of the ambiguous word, while we generate substitutes for the first occurrence only. 
For the other two datasets ELMo LMs trained on WMT News and Wikipedia give comparable results, the former having slightly better maxARI on bts-rnc. ELMo LMs trained on Twitter perform slightly but consistently worse, and ULMFiT gives much worse results than other models. We suppose that bad performance of the ULMFiT model for WSI may be related to relatively small vocabulary (60K words compared to 1M in ELMo LMs), which may prevent generating substitutes that can discriminate different senses. 
Based on these results we have selected ELMo LMs trained on WMT News for all further experiments on bts-rnc and active-dict, and the ones trained on Wikipedia for wiki-wiki. It is also worth exploring combinations of different models trained on different corpora in the future work.

\subsection{Comparison of LM combination methods}
\begin{figure}[h]
\centering
\includegraphics[width=1.0\textwidth]{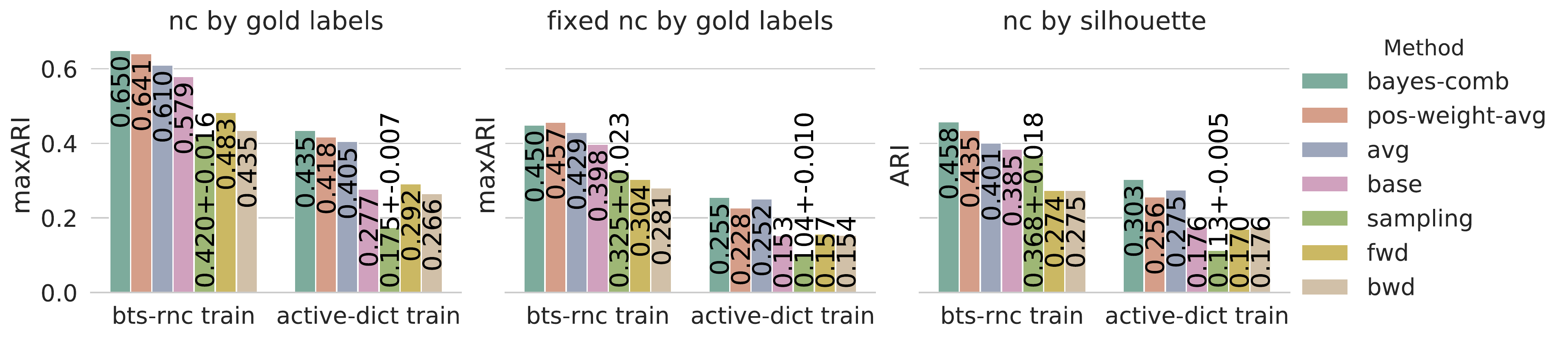}
\caption{Comparison of LM combination methods. \textit{Fixed nc} denotes using the same number of clusters for all words.}
\label{img:cmp_fixnc_msnc}
\end{figure}

Figure \ref{img:cmp_fixnc_msnc} shows WSI results on the train sets depending on forward and backward LM combination method compared to the baselines and unidirectional LMs alone. Bayesian combination is the best combination method on all datasets according to both maxARI and ARI (excluding maxARI on bts-rnc for the fixed number of clusters which is worse than individual number of clusters anyway). Surprisingly, the original method (\textbf{sampling}) often performs worse than unidirectional LMs. However, our deterministic modification (\textbf{base}) performs better or comparable to them and all proposed combination methods improve its results.

It is trivial that performance upper bound when selecting individual number of clusters per word (nc by gold labels) is always better than using the same number of clusters for all words (fixed nc by gold labels). However, we would like to emphasize the large difference. Surprisingly, selecting individual number of clusters based on silhouette score (nc by silhouette), which doesn't exploit gold labels, often gives better or similar results to the upper bound for the fixed number of clusters. Unfortunately, the margin between ARI and maxARI is still large so it worth experimenting with other methods of selecting individual number of clusters.

\subsection{Comparison to the previous best results}
Table \ref{tab:main-results} compares our baselines and best methods to the previous best methods according to the official leaderboard. The best previous results are improved on bts-rnc and active-dict by a large margin (which is especially large on bts-rnc presumably due to longer contexts which make neural LMs generating better substitutes). The results of our baseline method is also better than previous results on bts-rnc, but little worse on active-dict, which underlines the benefits of combining forward and backward LMs properly, especially when contexts are short. For both the baseline and the best vectorization method selecting the number of clusters based on silhouette score works significantly better than using fixed number of clusters selected on the train sets.

\begin{table}[]
\centering
\scriptsize
\begin{tabular}{|l|c|c|c|c|c|c|}
\hline
\multicolumn{1}{|c|}{\multirow{3}{*}{\textbf{Model}}} & \multicolumn{2}{c|}{\textbf{bts-rnc}} & \multicolumn{2}{c|}{\textbf{wiki-wiki}} & \multicolumn{2}{c|}{\textbf{active-dict}} \\ \cline{2-7} 
\multicolumn{1}{|c|}{} & \multicolumn{2}{c|}{\textbf{Test ARI}} & \multicolumn{2}{c|}{\textbf{Test ARI}} & \multicolumn{2}{c|}{\textbf{Test ARI}} \\ \cline{2-7} 
\multicolumn{1}{|c|}{} & \textbf{Public} & \textbf{Private} & \textbf{Public} & \textbf{Private} & \textbf{Public} & \textbf{Private} \\ \hline
bayes-comb-silnc & \textbf{0.502} & \textbf{0.451} & 0.651 & 0.616 & \textbf{0.331} & \textbf{0.298} \\ \hline
bayes-comb-fixnc & 0.464 & 0.438 & 0.651 & \textbf{0.682} & 0.304 & 0.260 \\ \hline
base-silnc & 0.365 & 0.362 & 0.651 & 0.646 & 0.202 & 0.162 \\ \hline
base-fixnc & 0.328 & 0.298 & 0.651 & 0.394 & 0.143 & 0.141 \\ \hline
\hline
post-competition improvement & - & - & - & - & \textbf{0.307} & 0.234 \\ \hline
competition best result & \textbf{0.351} & \textbf{0.338} & \textbf{1.0} & \textbf{0.962} & 0.264 & \textbf{0.248} \\ \hline
competition 2nd best result & 0.281 & 0.281 & 1.0 & 0.659 & 0.236 & 0.227 \\ \hline
\end{tabular}
\caption{Comparison with previous best results. Selecting the number of clusters individually using silhouette score (silnc) or as a hyperparameter on train (fixnc).}
\label{tab:main-results}
\end{table}

In the public wiki-wiki test set there are only 2 words, one of them was clustered perfectly by all methods and another has only one sense while our methods split it into two clusters. The public test set contains 4 words for which our results are comparable to the competition 2nd best results but are much worse than the best results. The analysis of these results has revealed that such a large gap is mostly due to suboptimal number of clusters selected by our methods on wiki-wiki. Using our vectorization and clustering but with the same number of clusters as in the best submission improves our ARI on the private test set to 0.89 while maxARI is 0.95. Also it is possible that substitutes-based methods are suboptimal for homonyms because unrelated senses are likely to be accompanied by unrelated context words, so context clustering approaches may perform better in this scenario. However, experiments on larger datasets consisting of homonyms are needed to check this hypothesis.

\begin{table}[]
\centering
\scriptsize
\begin{tabular}{|l|c|c|c|c|c|c|c|c|}
\hline
\multicolumn{1}{|c|}{\multirow{2}{*}{\textbf{Model}}} & \multicolumn{4}{c|}{\textbf{Public Test}} & \multicolumn{4}{c|}{\textbf{Private Test}} \\ \cline{2-9} 
\multicolumn{1}{|c|}{} & \textbf{F-Sc} & \textbf{V-M} & \textbf{AVG} & \textbf{\#Cl} & \textbf{F-Sc} & \textbf{V-M} & \textbf{AVG} & \textbf{\#Cl} \\ \hline
bayes-comb-silnc & 0.795 & \textbf{0.454} & \textbf{0.601} & 4.0 & \textbf{0.776} & \textbf{0.432} & \textbf{0.579} & 3.8 \\ \hline
bayes-comb-fixnc & \textbf{0.798} & 0.404 & 0.568 & 4.0 & 0.772 & 0.421 & 0.570 & 4.0 \\ \hline
base-silnc & 0.721 & 0.351 & 0.503 & 2.9 & 0.702 & 0.363 & 0.505 & 2.9 \\ \hline
base-fixnc & 0.774 & 0.271 & 0.458 & 4.0 & 0.756 & 0.294 & 0.471 & 4.0 \\ \hline
\hline
post-competition improvement & - & - & - & - & - & - & - & - \\ \hline
competition best result & 0.731 & 0.271 & 0.445 & 2.4 & 0.710 & 0.3 & 0.462 & 2.3 \\ \hline
competition 2nd best result & 0.692 & 0.288 & 0.446 & 10.0 & 0.683 & 0.298 & 0.451 & 10.0 \\ \hline
\hline
1 cluster per word & 0.764 & 0.0 & 0.0 & 1.0 & 0.726 & 0.0 & 0.0 & 1.0 \\ \hline
1 cluster per instance & 0.0 & 0.221 & 0.004 & 130.6 & 0.0 & 0.244 & 0.005 & 127.5 \\ \hline
\end{tabular}
\caption{Semeval 2010 metrics on bts-rnc.}
\label{tab:s2010_metrics_bts_rnc}
\end{table}

To provide better comparison, in tables~\ref{tab:s2010_metrics_bts_rnc},\ref{tab:s2010_metrics_active_dict} we report the metrics from SemEval 2010 Task 14 which is a similar competition for English. We have downloaded the previous best submissions from the RUSSE 2018 leaderboard and used the official SemEval 2010 evaluation script to calculate paired F-score (F-Sc) and V-measure (V-M). The results of two primitive baselines, one placing all examples into a single cluster and another allocating a separate cluster for each example, are calculated to show that both F-score and V-measure are highly biased towards small or large number of clusters and rather useless in isolation. Thus, we additionally report their geometric mean (AVG) and the average number of clusters per word (\#Cl).
On bts-rnc our methods outperform all previously best methods according to all metrics. On active-dict F-score of our methods is little worse, but V-measure and, most importantly, AVG are better. Worse F-Score can be explained by the larger number of clusters our methods produce. Appendix~\ref{appendix_analysis} provides more detailed analysis of pros and cons of our approach compared to the previous best submissions.


\begin{table}[]
\centering
\scriptsize
\begin{tabular}{|l|c|c|c|c|c|c|c|c|}
\hline
\multicolumn{1}{|c|}{\multirow{2}{*}{\textbf{Model}}} & \multicolumn{4}{c|}{\textbf{Public Test}} & \multicolumn{4}{c|}{\textbf{Private Test}} \\ \cline{2-9} 
\multicolumn{1}{|c|}{} & \textbf{F-Sc} & \textbf{V-M} & \textbf{AVG} & \textbf{\#Cl} & \textbf{F-Sc} & \textbf{V-M} & \textbf{AVG} & \textbf{\#Cl} \\ \hline
bayes-comb-silnc & 0.484 & 0.538 & \textbf{0.511} & 5.2 & 0.459 & 0.505 & \textbf{0.482} & 5.4 \\ \hline
bayes-comb-fixnc & 0.453 & 0.527 & 0.489 & 6.0 & 0.421 & 0.479 & 0.449 & 6.0 \\ \hline
base-silnc & 0.401 & 0.395 & 0.398 & 4.6 & 0.362 & 0.365 & 0.363 & 5.1 \\ \hline
base-fixnc & 0.349 & 0.367 & 0.358 & 5.0 & 0.351 & 0.352 & 0.351 & 5.0 \\ \hline
\hline
post-competition improvement & \textbf{0.513} & 0.451 & 0.481 & 3.0 & 0.464 & 0.389 & 0.425 & 3.0 \\ \hline
competition best result & 0.489 & 0.411 & 0.445 & 3.2 & 0.467 & 0.392 & 0.428 & 3.4 \\ \hline
competition 2nd best result & 0.468 & 0.377 & 0.420 & 3.0 & \textbf{0.468} & 0.371 & 0.417 & 3.0 \\ \hline
\hline
1 cluster per word & 0.433 & 0.0 & 0.0 & 1.0 & 0.437 & 0.0 & 0.0 & 1.0 \\ \hline
1 cluster per instance & 0.0 & \textbf{0.55} & 0.014 & 21.9 & 0.0 & \textbf{0.543} & 0.010 & 22.4 \\ \hline
\end{tabular}
\caption{Semeval 2010 metrics on active-dict.}
\label{tab:s2010_metrics_active_dict}
\end{table}

\subsection{Analysis of the results}
\label{appendix_analysis}
In this section we answer two questions: do we select the number of clusters better and can we cluster a target word occurrences into given number of clusters better than the previous best method? We compare our best submissions to the previous best submissions according to the private test ARI on bts-rnc and active-dict. 

To answer the first question, we calculated the mean squared error (MSE) between the number of clusters in each submission and the true number of senses. Our method estimates the number of senses much worse (MSE is 3.41 versus 1.65 on bts-rnc and 8.48 versus 1.20 on active-dict), usually returning larger number of clusters. But is the optimal number of clusters equal to the true number of senses? In our case the data contains outliers (i.e. points which are far from all other points due to non-informative context or other vectorization problems). Thus, the number of clusters should be larger than the true number of senses. Otherwise, some senses will be merged with other senses while outliers will occupy their clusters. The results below and Appendix~\ref{appendix_nc} provide some evidences that our method better estimates the optimal number of clusters.

\begin{figure}[h]
\centering
\includegraphics[width=1.0\textwidth]{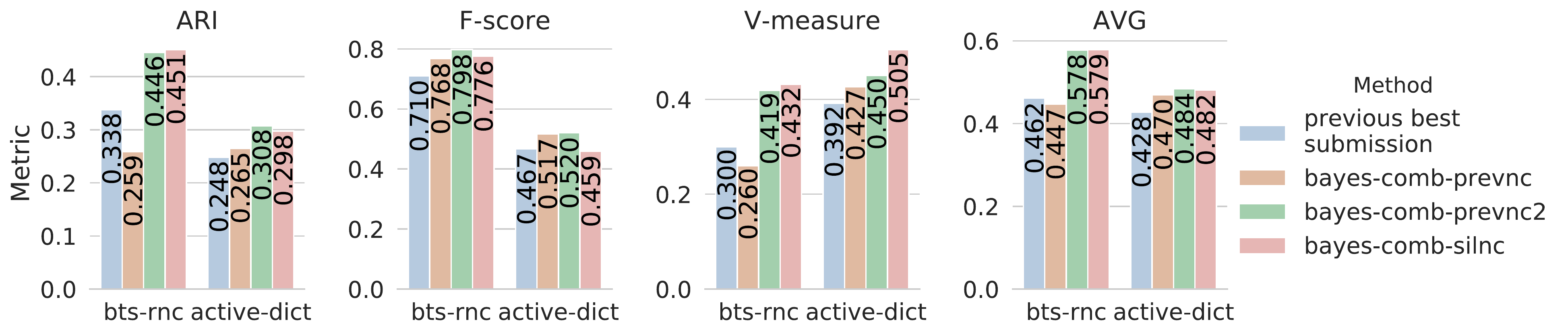}
\caption{Comparison of our and previous best submissions with different and equal number of clusters.}
\label{img:nc_by_best_result}
\end{figure}

It is hard to make comparison of vectorization and clustering algorithms separately, since we don't have vectors and cannot vary the number of clusters for the previous methods, only clustering results are available. One thing we can do is clustering examples of i-th word into the number of clusters $P_i$ taken from the previous best submission, but using our vectors and clustering algorithm. Then the difference in performance cannot be due to the difference in the number of clusters selected. This is achieved by two techniques. The first one (bayes-comb-prevnc) simply sets the number of clusters equal to $P_i$ in our agglomerative clustering, which is suboptimal due to outliers. Figure~\ref{img:nc_by_best_result} shows that this works much worse than the number of clusters selected using silhouette score according to all metrics (except F-score on active-dict). Compared to the previous best submissions the results are mixed. Another technique (bayes-comb-prevnc2) first clusters the occurrences of each word into the number of clusters $S_i$ selected by silhouette score. If $S_i>P_i$, then we leave $P_i$ largest clusters intact and distribute all other examples among them by moving each example to the nearest cluster. Otherwise, we simply cluster examples into $P_i$ clusters like before. This results in much better clustering compared to the previous best submissions given the same number of clusters according to all metrics. To conclude, compared to the previous best submissions our approach has selected the number of clusters that is better according to several evaluation metrics for our vectorization, but further from the true number of senses. It can also cluster the datasets better when the number of clusters is taken from another submissions and is not optimal given our vectorization. 

\section{Conclusion}
Bidirectional neural LMs are a powerful instrument for different tasks including substitutes generation for WSI, but some tricks should be used to apply them to this task properly. We have proposed and compared several methods of combining forward and backward LMs for better substitutes generation. Also we have proposed a technique for selecting individual number of clusters per word and this improved results even further. We improved previous best results on two datasets from RUSSE 2018 WSI shared task for the Russian language by a large margin. Finally, we have compared several Russian LMs regarding their performance for WSI.

\section{Acknowledgements}
We are grateful to Dima Lipin, Artem Grachev and Alex Nevidomsky for their valuable help.

\bibliographystyle{splncs03}
\bibliography{aist}

\begin{thebibliography}{10}
\providecommand{\url}[1]{\texttt{#1}}
\providecommand{\urlprefix}{URL }

\bibitem{alagic_leveraging}
Alagi{\'c}, D., {\v{S}}najder, J., Pad{\'o}, S.: Leveraging lexical substitutes
  for unsupervised word sense induction. In: Thirty-Second AAAI Conference on
  Artificial Intelligence (2018)

\bibitem{amplayo2019granularity}
Amplayo, R.K., won Hwang, S., Song, M.: Autosense model for word sense
  induction. In: AAAI (2019)

\bibitem{amrami_goldberg}
Amrami, A., Goldberg, Y.: Word sense induction with neural bi{LM} and symmetric
  patterns. In: Proceedings of the 2018 Conference on Empirical Methods in
  Natural Language Processing. pp. 4860--4867. Association for Computational
  Linguistics, Brussels, Belgium (2018),
  \url{https://www.aclweb.org/anthology/D18-1523}

\bibitem{apresjan2011active}
Apresjan, V.: Active dictionary of the russian language: theory and practice.
  Meaning-Text Theory  2011,  13--24 (2011)

\bibitem{RUSSE_Arefyev}
Arefyev, N., Ermolaev, P., Panchenko, A.: How much does a word weigh? weighting
  word embeddings for word sense induction. In: Computational Linguistics and
  Intellectual Technologies: Papers from the Annual International Conference
  ``Dialogue''. pp. 68--84. RSUH, Moscow, Russia (2018)

\bibitem{bartunov2015breaking}
Bartunov, S., Kondrashkin, D., Osokin, A., Vetrov, D.: Breaking sticks and
  ambiguities with adaptive skip-gram. In: Proceedings of the International
  Conference on Artificial Intelligence and Statistics (AISTATS) (2016)

\bibitem{baskaya}
Baskaya, O., Sert, E., Cirik, V., Yuret, D.: Ai-ku: Using substitute vectors
  and co-occurrence modeling for word sense induction and disambiguation. In:
  Second Joint Conference on Lexical and Computational Semantics (* SEM),
  Volume 2: Proceedings of the Seventh International Workshop on Semantic
  Evaluation (SemEval 2013). vol.~2, pp. 300--306 (2013)

\bibitem{bert}
Devlin, J., Chang, M.W., Lee, K., Toutanova, K.: Bert: Pre-training of deep
  bidirectional transformers for language understanding. arXiv preprint
  arXiv:1810.04805  (2018)

\bibitem{Hope2013}
Hope, D., Keller, B.: {UoS: A Graph-Based System for Graded Word Sense
  Induction}. In: Second Joint Conference on Lexical and Computational
  Semantics (*SEM), Volume 2: Proceedings of the Seventh International Workshop
  on Semantic Evaluation (SemEval 2013). pp. 689--694. No.~1, Atlanta, Georgia,
  USA (2013), \url{http://www.aclweb.org/anthology/S13-2113}

\bibitem{ulmfit}
Howard, J., Ruder, S.: Universal language model fine-tuning for text
  classification. In: Proceedings of the 56th Annual Meeting of the Association
  for Computational Linguistics (Volume 1: Long Papers). pp. 328--339.
  Association for Computational Linguistics, Melbourne, Australia (2018),
  \url{https://www.aclweb.org/anthology/P18-1031}

\bibitem{jurgens}
Jurgens, D., Klapaftis, I.: Semeval-2013 task 13: Word sense induction for
  graded and non-graded senses. In: Second Joint Conference on Lexical and
  Computational Semantics (* SEM), Volume 2: Proceedings of the Seventh
  International Workshop on Semantic Evaluation (SemEval 2013). vol.~2, pp.
  290--299 (2013)

\bibitem{RUSSE_Kutuzov}
Kutuzov, A.: Russian word sense induction by clustering averaged word
  embeddings. CoRR  abs/1805.02258 (2018),
  \url{http://arxiv.org/abs/1805.02258}

\bibitem{Lau2013}
Lau, J.H., Cook, P., Baldwin, T.: {unimelb: Topic Modelling-based Word Sense
  Induction}. In: Second Joint Conference on Lexical and Computational
  Semantics (*SEM): SemEval 2013). vol.~2, pp. 307--311. Atlanta, Georgia, USA
  (2013), \url{http://www.aclweb.org/anthology/S13-2051}

\bibitem{manandhar}
Manandhar, S., Klapaftis, I.P., Dligach, D., Pradhan, S.S.: Semeval-2010 task
  14: Word sense induction \& disambiguation. In: Proceedings of the 5th
  international workshop on semantic evaluation. pp. 63--68. Association for
  Computational Linguistics (2010)

\bibitem{melamud_context2vec}
Melamud, O., Goldberger, J., Dagan, I.: context2vec: Learning generic context
  embedding with bidirectional lstm. In: Proceedings of The 20th SIGNLL
  Conference on Computational Natural Language Learning. pp. 51--61 (2016)

\bibitem{panchenko_2018}
Panchenko, A., Lopukhina, A., Ustalov, D., Lopukhin, K., Arefyev, N., Leontyev,
  A., Loukachevitch, N.: {RUSSE'2018: A Shared Task on Word Sense Induction for
  the Russian Language}. In: Computational Linguistics and Intellectual
  Technologies: Papers from the Annual International Conference ``Dialogue''.
  pp. 547--564. RSUH, Moscow, Russia (2018),
  \url{http://www.dialog-21.ru/media/4324/panchenkoa.pdf}

\bibitem{elmo_model}
Peters, M.E., Neumann, M., Iyyer, M., Gardner, M., Clark, C., Lee, K.,
  Zettlemoyer, L.: Deep contextualized word representations. In: Proc. of NAACL
  (2018)

\bibitem{RUSSE_Transformer}
Struyanskiy, O., Arefyev, N.: Neural networks with attention for word sense
  induction. In: Supplementary Proceedings of the Seventh International
  Conference on Analysis of Images, Social Networks and Texts {(AIST} 2018),
  Moscow, Russia, July 5 - 7, 2018. pp. 208--213 (2018),
  \url{http://ceur-ws.org/Vol-2268/paper23.pdf}

\bibitem{self_attention}
Tang, G., M{\"u}ller, M., Rios, A., Sennrich, R.: Why self-attention? a
  targeted evaluation of neural machine translation architectures. arXiv
  preprint arXiv:1808.08946  (2018)

\bibitem{attention}
Vaswani, A., Shazeer, N., Parmar, N., Uszkoreit, J., Jones, L., Gomez, A.N.,
  Kaiser, {\L}., Polosukhin, I.: Attention is all you need. In: Advances in
  neural information processing systems. pp. 5998--6008 (2017)

\bibitem{veronis2004hyperlex}
V{\'e}ronis, J.: Hyperlex: lexical cartography for information retrieval.
  Computer Speech \& Language  18(3),  223--252 (2004)

\bibitem{bertmouth}
Wang, A., Cho, K.: {BERT} has a mouth, and it must speak: {BERT} as a markov
  random field language model. CoRR  abs/1902.04094 (2019),
  \url{http://arxiv.org/abs/1902.04094}

\bibitem{DBLP:journals/tacl/WangBGZY15}
Wang, J., Bansal, M., Gimpel, K., Ziebart, B.D., Yu, C.T.: A sense-topic model
  for word sense induction with unsupervised data enrichment. {TACL}  3,
  59--71 (2015)

\end{thebibliography}

\newpage
\appendix
\section{Examples of substitutes generated}
\label{appendix_examples}
Table~\ref{tab:discriminative-substitutes} provides examples of discriminative substitutes with their relative frequencies for each of two most frequent senses of several words. A substitute is called discriminative if it is frequently generated for one sense of an ambigusous word, but rarely for another. Formally, we take substitutes with the largest
$\frac{P(w|sense_1)}{P(w|sense_2)}$, where $P(w|sense_i)$ is estimated using add-one smoothing:
$$P(w|sense_i) = \frac{cnt(w|sense_i) + 1}{cnt(sense_i)+|vocab|}$$
Additionally, we leave only substitutes which were generated at lest 10 times for one of the senses.

\begin{table}[]
\centering
\scriptsize
\begin{tabular}{|c|c|c|}
\hline
\textbf{Балка} & \textbf{Штамп} & \textbf{Лавка} \\ \hline
\begin{tabular}[c]{@{}c@{}}Number of examples: 81\\ Sense1: Горизонтальный \\ опорный брус\end{tabular} & \begin{tabular}[c]{@{}c@{}}Number of examples: 45\\ Sense1: Рельефное устройство \\ для получения одинаковых \\ графических оттисков\end{tabular} & \begin{tabular}[c]{@{}c@{}}Number of examples: 67\\ Sense1: Скамья для \\ сидения или лежания\end{tabular} \\ \hline
\begin{tabular}[c]{@{}c@{}}перегородка 0.56/0.00\\ люстра 0.52/0.00\\ карниз 0.48/0.00\\ крышка 0.43/0.00\\ панель 0.41/0.00\\ козырёк 0.35/0.00\\ каркас 0.33/0.00\\ потолок 0.33/0.00\\ перекрытие 0.33/0.00\\ пластина 0.32/0.00\end{tabular} & \begin{tabular}[c]{@{}c@{}}справка 0.58/0.00\\ печать 0.56/0.00\\ подпись 0.53/0.00\\ пометка 0.51/0.00\\ бирка 0.44/0.00\\ талон 0.42/0.00\\ ксерокопия 0.42/0.00\\ выписка 0.42/0.00\\ прописка 0.40/0.00\\ бланк 0.40/0.00\end{tabular} & \begin{tabular}[c]{@{}c@{}}корточки 0.49/0.00\\ подушка 0.46/0.00\\ коврик 0.46/0.00\\ трибуна 0.39/0.00\\ простыня 0.37/0.00\\ одеяло 0.36/0.00\\ четвереньки 0.36/0.00\\ палуба 0.33/0.00\\ каталка 0.28/0.00\\ носилки 0.25/0.00\end{tabular} \\ \hline
\hline
\begin{tabular}[c]{@{}c@{}}Number of examples: 38\\ Sense2: Длинный и \\ широкий овраг\end{tabular} & \begin{tabular}[c]{@{}c@{}}Number of examples: 47\\ Sense2: Принятый образец, \\ которому следуют \\ без размышлений\end{tabular} & \begin{tabular}[c]{@{}c@{}}Number of examples: 82\\ Sense2: Небольшой \\ магазин\end{tabular} \\ \hline
\begin{tabular}[c]{@{}c@{}}деревня 0.01/0.47\\ степь 0.01/0.53\\ речушка 0.00/0.29\\ тайга 0.00/0.29\\ перевал 0.00/0.29\\ бухта 0.00/0.29\\ озеро 0.00/0.29\\ долина 0.00/0.34\\ речка 0.01/0.74\\ роща 0.00/0.42\end{tabular} & \begin{tabular}[c]{@{}c@{}}гений 0.02/0.26\\ стереотип 0.04/0.40\\ сюжет 0.02/0.28\\ миф 0.02/0.28\\ пафос 0.02/0.28\\ ритм 0.00/0.23\\ скучный 0.00/0.23\\ канон 0.00/0.26\\ стиль 0.00/0.28\\ жанр 0.00/0.30\end{tabular} & \begin{tabular}[c]{@{}c@{}}гостиница 0.00/0.28\\ закусочный 0.00/0.30\\ типография 0.00/0.33\\ касса 0.00/0.34\\ фабрика 0.00/0.34\\ супермаркет 0.00/0.35\\ бар 0.00/0.37\\ кофейня 0.00/0.37\\ магазин 0.00/0.44\\ аптека 0.00/0.48\end{tabular} \\ \hline
\end{tabular}
\caption{Discriminative substitutes for several words from bts-rnc train}
\label{tab:discriminative-substitutes}
\end{table}

Table~\ref{tab:list-substitutes} lists ten most probable substitutes according to the combined distribution and according to the forward and the backward LM distributions separately for several examples. Substitutes from unidirectional distributions are very sensitive to the position of the target word. When either left or right context doesn't contain enough information at least halve of the substitutes will be not related to the target word. Combined distribution provides more relevant substitutes.

\begin{table}[]
\centering
\scriptsize
\begin{tabular}{|c|c|c|}
\hline
\textbf{Bayes-comb} & \textbf{Base forward} & \textbf{Base backward} \\ \hline
\multicolumn{3}{|c|}{\begin{tabular}[c]{@{}c@{}}Нет , я по-прежнему проживаю в своей квартире , и в паспорте есть нужный \\ \textbf{штамп}. Просто сотни жителей в моем и соседних домах уже несколько \\ месяцев живут\end{tabular}} \\ \hline
\begin{tabular}[c]{@{}c@{}}штамп, этаж, номер, дом, ключ, \\ абзац, прочерк, подпункт, пункт, \\ пробел\end{tabular} & \begin{tabular}[c]{@{}c@{}}штамп, номер, пункт, \\ документ, знак\end{tabular} & \begin{tabular}[c]{@{}c@{}}ул, м., обл, \\ см, пр\end{tabular} \\ \hline
\hline
\multicolumn{3}{|c|}{\begin{tabular}[c]{@{}c@{}}Он был очень высок , наклонил голову , словно подпирая плечом \\ потолочную \textbf{балку}, посмотрел на Сьянову серьезными черными глазами .\end{tabular}} \\ \hline
\begin{tabular}[c]{@{}c@{}}занавеску, перегородку, ручку, \\ подушку, стенку, табуретку,\\ проволоку, стену, раму, плиту\end{tabular} & \begin{tabular}[c]{@{}c@{}}стенку, подушку,\\ перегородку, дверь,\\ стену\end{tabular} & \begin{tabular}[c]{@{}c@{}}увидел, помню,\\ оглянулся,\\ посмотрел, встал\end{tabular} \\ \hline
\hline
\multicolumn{3}{|c|}{\begin{tabular}[c]{@{}c@{}}Иногда туманным , осенним вечером он проходил вдоль \textbf{опушки} леса , \\ шурша омертвевшими листьями , подобрав длинную , черную рясу ;\end{tabular}} \\ \hline
\begin{tabular}[c]{@{}c@{}}опушки, кромки, заснеженного, \\ живописного, посреди, \\ соснового, цветущего, чащи, \\ тропического, стога\end{tabular} & \begin{tabular}[c]{@{}c@{}}забора, берега, \\ стен, реки, \\ кромки\end{tabular} & \begin{tabular}[c]{@{}c@{}}посреди, глубь, \\ опушке, вдоль, \\ вглубь\end{tabular} \\ \hline
\hline
\multicolumn{3}{|c|}{\begin{tabular}[c]{@{}c@{}}Однажды , когда страховой агент заполнял мой \textbf{полис} , он допустил \\ ошибку и написал меньшее количество лошадиных сил .\end{tabular}} \\ \hline
\begin{tabular}[c]{@{}c@{}}бланк, талон, формуляр, полис, \\ анкету, талончик, бак, протокол, \\ водительский, профиль\end{tabular} & \begin{tabular}[c]{@{}c@{}}бланк, номер, \\ полис, паспорт, \\ анкету\end{tabular} & \begin{tabular}[c]{@{}c@{}}видимо, вероятно, \\ возможно, значит, \\ кажется\end{tabular} \\ \hline
\hline
\multicolumn{3}{|c|}{\begin{tabular}[c]{@{}c@{}}заранее старается выговорить для себя немало льгот . Так , например , \\ у него остается \textbf{пост} почетного президента ФХР , солидная \\ пенсия ( около 100 тысяч рублей в месяц ) , оплачиваемая\end{tabular}} \\ \hline
\begin{tabular}[c]{@{}c@{}}стипендия, привилегия, оклад, \\ виза, зарплата, диплом, значок, \\ вакансия, лицензия, выслуга\end{tabular} & \begin{tabular}[c]{@{}c@{}}право, возможность, \\ шанс, квартира, \\ масса\end{tabular} & \begin{tabular}[c]{@{}c@{}}звание, приз, \\ должность, \\ стипендия, пост\end{tabular} \\ \hline
\end{tabular}
\caption{Substitutes generated for randomly selected examples.}
\label{tab:list-substitutes}
\end{table}

\section{The number of clusters selected}
\label{appendix_nc}
\begin{figure}[h]
\centering
\includegraphics[width=1.0\textwidth]{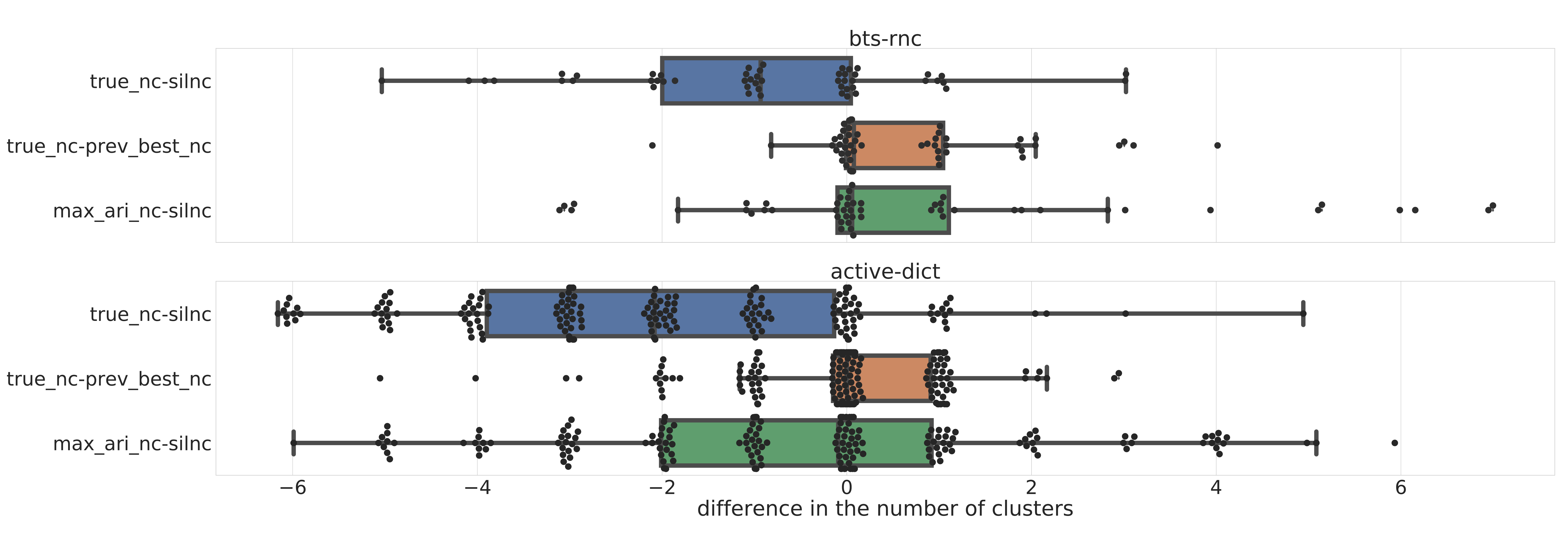}
\caption{Comparison of the number of clusters in our (silnc) and previous best submissions (prev\_best\_nc) with the true number of senses (true\_nc) and the optimal number of clusters (max\_ari\_nc).}
\label{img:nc_boxplot}
\end{figure}

Figure~\ref{img:nc_boxplot} plots distributions of the differences between the true number of senses, the number of clusters in submissions and the optimal number of clusters. Silhouette score gives the number of clusters, which is usually larger than the number of senses, but is near the optimum with respect to ARI and given our vectors. The previous best submissions better estimate the true number of senses.

\section{Hyperparameters}

Table \ref{tab:hyperparameters} shows the selected hyperparameters for the methods described in section~\ref{sec:approach}. For bts-rnc and active-dict datasets hyperparameters were selected using grid search on corresponding train sets. For wiki-wiki we used the hyperparameters from bts-rnc due to very small size of wiki-wiki train set. We selected the following hyperparameters. 
\begin{enumerate}
\item  \textbf{Add bias} (True/False). Ignoring bias in the softmax layer of the LM was proposed by \cite{amrami_goldberg} to improve substitutes, because adding bias results in prediction of frequency words instead of rare but relevant substitutes.
\item \textbf{Normalize output embeddings} (True/False). Similarly to ignoring bias, this may result in prediction of more relevant substitutes.
\item \textbf{K} (10-400). The number of substitutes from each distribution.
\item \textbf{Exclude Target} (True/False). We want the substitutes for different senses of the target word to be non-overlapping. Thus, it may be beneficial to exclude the target word from the substitutes.
\item \textbf{TFIDF} (True/False). Applying TFIDF transformation to bag-of-words vectors of substitutes sometimes improve performance.
\item \textbf{S} (=20). The number of representatives for each example. It didn't affect the performance so we use the value from \cite{amrami_goldberg}. 
\item \textbf{L} (4-30). The number of substitutes to sample from top K predictions.
\item \textbf{z} (1.0-3.0). The parameter of Zipf distribution.
\item \boldmath{$\beta$} (0.1-0.5). Relative length of the left or the right context after which the discounting of the corresponding LM begins.
\end{enumerate}

\begin{table}[h]
\centering
\scriptsize
\begin{tabular}{|l|c|c|c|c|c|c|c|c|c|}
\hline
\multicolumn{1}{|c|}{\textbf{Method}} & \textbf{Add bias} & \textbf{\begin{tabular}[c]{@{}c@{}}Normalize \\ output \\ embeddings\end{tabular}} & \textbf{K} & \textbf{\begin{tabular}[c]{@{}c@{}}Exclude\\ Target\end{tabular}} & \textbf{TF-IDF} & \textbf{S} & \textbf{L} & \multicolumn{1}{c|}{\textbf{z}} & \multicolumn{1}{c|}{\boldmath{$\beta$}} \\ \hline
\multicolumn{10}{|c|}{\textbf{bts-rnc}} \\ \hline
\textbf{bayes comb} & False & False & 200 & True & False & - & - & 2.0 & - \\ \hline
\textbf{pos weight avg} & False & False & 200 & True & False & - & - & - & 0.1 \\ \hline
\textbf{avg} & False & False & 150 & True & False & - & - & - & - \\ \hline
\textbf{base} & False & False & 200 & True & False & - & - & - & - \\ \hline
\textbf{sampling} & False & False & 200 & True & True & 20 & 15 & - & - \\ \hline
\textbf{forward} & False & True & 150 & True & False & - & - & - & - \\ \hline
\textbf{backward} & False & False & 300 & True & False & - & - & - & - \\ \hline
\multicolumn{10}{|c|}{\textbf{active-dict}} \\ \hline
\textbf{bayes comb} & False & False & 200 & True & True & - & - & 2.0 & - \\ \hline
\textbf{pos weight avg} & False & False & 200 & True & True & - & - & - & 0.1 \\ \hline
\textbf{avg} & False & False & 150 & True & True & - & - & - & - \\ \hline
\textbf{base} & False & False & 200 & True & True & - & - & - & - \\ \hline
\textbf{sampling} & False & False & 200 & True & True & 20 & 10 & - & - \\ \hline
\textbf{forward} & False & True & 100 & True & True & - & - & - & - \\ \hline
\textbf{backward} & False & True & 300 & True & True & - & - & - & - \\ \hline
\end{tabular}
\caption{Selected hyperparameters}
\label{tab:hyperparameters}
\end{table}

\end{document}